\documentclass[10pt,twocolumn,letterpaper]{article}
\usepackage[pagenumbers]{cvpr}
\usepackage{algorithmic}
\usepackage{algorithm}
\usepackage{array}
\usepackage{textcomp}
\usepackage{verbatim}
\usepackage{multirow}
\usepackage{threeparttable}
\usepackage{makecell}
\usepackage{dsfont}
\usepackage{pifont}
\newcommand{\cmark}{\ding{51}}
\newcommand{\xmark}{\ding{55}}

\definecolor{cvprblue}{rgb}{0.21,0.49,0.74}
\usepackage[breaklinks,colorlinks,allcolors=cvprblue]{hyperref}
\usepackage[capitalize]{cleveref}
\title{Let Samples Sing in Chorus: \\
Mitigating Spurious Correlations via Cross-view Consensus}
\author{
Weiwei Li$^{1}$ \quad Junzhuo Liu$^{1}$ \quad Tong Chu$^{1}$ \quad Hengfu Yu$^{1}$ \\
Yinjie Lei$^{2}$ \quad Wen Li$^{1}$\thanks{Corresponding author.} \\[4pt]
{\normalsize $^{1}$University of Electronic Science and Technology of China} \\
{\normalsize $^{2}$Sichuan University} \\[4pt]
{\small\texttt{davelee.uestc@gmail.com, junzhuo.cs@gmail.com, uestcchutong@gmail.com}} \\
{\small\texttt{hfyu@std.uestc.edu.cn, yinjie@scu.edu.cn, liwenbnu@gmail.com}}
}
\hypersetup{pdftitle={Let Samples Sing in Chorus:
Mitigating Spurious Correlations via Cross-view Consensus},pdfauthor={Weiwei Li, Junzhuo Liu, Tong Chu, Hengfu Yu, Yinjie Lei, Wen Li},pdfsubject={Spurious correlations and group-robust learning},pdfkeywords={CHORD, spurious correlations, debiasing, worst-group accuracy}}
\begin{document}
\maketitle
\begin{abstract}
    Deep models often rely on spurious correlations that fail on underrepresented groups. Existing methods either correct human-specified bias attributes or infer unknown biases from the learned representation space. However, annotated attributes may differ from the shortcuts actually used by a model, while a single representation provides incomplete and potentially noisy evidence. Our key insight is that a bias obscured in one feature space may be clearer in another. In this work, we propose CHORD (Cross-view Harmonic Outlier Reversal for Debiasing), a multi-view framework that allows samples to "sing in chorus" across diverse representation spaces to achieve robust cross-view consensus. CHORD aggregates aggregates within-class partitions in each view, which is a proxy for specifying candidate biases,  to debias the target representation and classifier without bias supervision. Experiments on seven benchmarks across vision, language, speech, and medical imaging show consistent improvements in worst-group accuracy.
    Code and checkpoints are available at \url{https://github.com/davelee-uestc/chord_debiasing_public}.

\end{abstract}

\section{Introduction}
\label{sec:intro}

Deep neural networks often exploit spurious features that are highly correlated with the label for most training samples but fail on underrepresented groups \cite{geirhos2020shortcut,sagawa2020investigation}. 
For example, a biased model may recognize birds by their backgrounds rather than their appearance, performing well on majority groups such as waterbirds on water, but misclassifying bias-conflicting samples in minority groups such as waterbirds on land. Similarly, it may diagnose diseases from medical devices rather than the pathology itself \cite{bissoto2019constructing,zech2018variable,oakden2020hidden}.
Many previous methods address this problem using predefined spurious attributes or group annotations from human experts \cite{sagawa2019distributionally,kirichenko2022last,tsirigotis2024group}. 
However, annotation is costly, and the annotated attribute may not be the one the model actually uses. 
Models can favor features that differ from human expectations due to their architectures \cite{geirhos2018imagenet} or even random initializations \cite{allen2023towards}. Removing a human-annotated spurious attribute therefore does not guarantee that the model bias is removed.

Recent methods instead discover bias from the model to be debiased (the target model) itself due to its model-dependent nature. Existing methods use signals such as errors \cite{liu2021just}, losses \cite{nam2020learning}, and gradients \cite{creager21a} to identify bias-related samples. However, these signals can mix with label noise or intrinsic task difficulty, making it difficult to distinguish such effects from failures caused by spurious correlations. 
Representation-based methods instead seek readily available spurious features that models tend to learn, but require assumptions about the target geometry: GEORGE relies on separable subclasses \cite{sohoni2020no}. NSF relies on strong spurious features producing identifiable within-class deviations \cite{li2025nsf}. Such methods keep discovery tied to the target model, which is constrained by the its geometry. When a spurious feature is entangled with semantic variation or artifacts of geometry, the pattern found is a noisy estimate of the bias: it may still cover most bias-conflicting samples, yet they are diluted by unrelated ones. As in empirical risk minimization (ERM), the corrective signal can remain too weak to counter the majority pattern, leading to poor worst-group accuracy (WGA), the lowest accuracy across groups. 
Our starting point is simple: a noisy candidate bias pattern in one representation may be denoised in another. Models trained with different objectives organize the same samples differently and emphasize different attributes \cite{geirhos2018imagenet,hermannfoundations}. A spurious attribute that is difficult to disentangle from noisy cofounder in the target representation may therefore produce a clearer within-class division in another representation, which we call a view. This can potentially provide less noisy candidates for correcting the target model. Because all views observe the same samples, the candidate can be transferred through sample identities, without aligning feature spaces or naming the underlying attribute. Bias discovery can thus benefit from structures that are more visible outside the target representation.

Based on this insight, we propose Cross-view Harmonic Outlier Reversal for Debiasing (CHORD), a multi-view debiasing framework that requires no group annotations. Unlike prior methods constrained by the target geometry, CHORD decouples where candidates are proposed from where they are validated: candidates come from several views of the same samples, while validation stays in the target space. It proceeds in three phases: each view proposes a candidate bias, the target representation validates it, and the evidence from all views are aggregated for debiasing. Two design choices make this possible. First, CHORD represents each candidate bias as a within-class partition, allowing views to communicate their hypotheses through sample assignments without aligning feature spaces or specifying bias attributes. Second, CHORD filters views by testing every partition in the target space, since an auxiliary view, like a human annotator, may highlight attributes that are not the spurious cue to remove from the target features. The test uses a constrained transformation to suppress feature differences between the two subsets and keeps samples whose nearest-centroid assignments return to their labeled class. These reversal samples provide correction evidence from the partition. Each view decides where to look, while the target representation determines what views to keep and what samples contribute to debiasing. The evidence obtained from all views is aggregated into soft sampling weights for retraining the classifier head. Auxiliary views are frozen during training, and discarded at inference.

We evaluate CHORD on seven benchmarks spanning vision, language, and speech, with multiple architectures. On the four standard benchmarks, CHORD achieves the highest WGA among the compared methods without bias supervision. We further explore constructing views from heterogeneous modalities in the Discussion section.

Our contributions are as follows.
\begin{itemize}
\item We propose CHORD, a multi-view debiasing framework that decouples where biases are proposed from where they are tested. This allows bias discoverying on multiple representations, relaxing reliance on the target geometry.

    \item We represent candidate biases as within-class partitions, providing a transferable proxy that connects different representation spaces through sample assignments without feature alignment or predefined bias attributes. 

    \item We evaluate CHORD on seven benchmarks with multiple target architectures, showing highest WGA among competing methods without bias supervision.
\end{itemize}

\section{Related Work}
\label{sec:related_work}

\textbf{Mitigating Spurious Correlations.}
Spurious correlations allow models to exploit dataset-specific shortcuts that fail under distribution shifts \cite{geirhos2020shortcut,torralba2011unbiased}.
Existing methods broadly follow two routes: specifying bias information or inferring it through assumptions about learning behavior.
The former includes group-supervised robust optimization \cite{sagawa2019distributionally} and estimating missing spurious-attribute labels from limited annotations \cite{nam2022spread}.
Prior knowledge also guides interventions against texture dependence \cite{geirhos2018imagenet,he2023shift,MishraSBACS0022}, counterfactual image generation \cite{sauer2021counterfactual}, and methods that use language models to improve robustness \cite{adila2024zero}.
However, annotations are costly, and the specified attributes may not cover the shortcuts used by a particular model; suppressing one shortcut can even amplify another \cite{li2023whac}.
The other route reduces dependence on explicit bias labels by constructing deliberately biased auxiliary models \cite{bahng2020learning,nam2020learning} or exploiting simplicity bias \cite{tiwari2023overcoming}.
Initial-model errors or confidence \cite{liu2021just,qiu2023simple} and pseudo-bias labeling \cite{hu2023echoes} provide signals for emphasizing bias-conflicting samples.
Data-level approaches instead create or expose alternative training signals through bias-tailored swapping \cite{kim2021biaswap}, bias-adversarial augmentation \cite{lim2023biasadv}, or masking discriminative regions \cite{asgari2022masktune}.
Although these interventions differ, their effectiveness depends on how well the supplied or inferred bias matches the target model’s tendency to learn spurious features.
Learning-based proxies may also reflect ordinary task difficulty or noise, motivating bias discovery grounded in the target model without predefined bias attributes or group annotations.

\textbf{Representation-Based Bias Discovery and Mitigation.}
Learned representations support both lightweight correction and direct bias analysis.
DFR \cite{kirichenko2022last} and AFR \cite{qiu2023simple} exploit useful information retained by biased encoders through last-layer retraining, while contrastive objectives reshape feature relationships across bias conditions \cite{zhang2022correct,hong2021unbiased}.
Information-theoretic regularization also discourages shortcut dependence in dense prediction \cite{chuah2022itsa,chuah2023information}.
For discovery, clustering-derived pseudo-attributes guide debiased learning \cite{seo2022unsupervised}, and FACTS \cite{yenamandra2023facts} amplifies correlations before discovering bias-conflicting slices in feature space.
NSF \cite{li2025nsf} uses within-class geometry; ShortcutProbe \cite{zheng2025shortcutprobe} identifies latent prediction shortcuts; and Evidential Alignment \cite{ye2025evidential} uses uncertainty and evidence calibration to characterize and mitigate model bias.
Related approaches train diverse predictive hypotheses to explore alternatives to shortcut solutions \cite{teney2022evading,lee2023diversify}.
Within this literature, representation-based bias discovery remains sensitive to its own geometry.
Hermann et al.\ \cite{hermannfoundations} show that the features a model uses depend on both their predictive value and their availability, where availability refers to how easily they can be extracted from inputs.
This motivates considering differences in attribute visibility across representations: variation that is insufficiently prominent for discovery in the target space may be more accessible in another representation.
CHORD uses pretrained and geometry-enhanced views to supply candidate partitions, reducing reliance on the target geometry that spurious variation being sufficiently prominent.
The candidates are evaluated in the target space, so the additional views broaden the sources of evidence while keeping bias discovery tied to the model being corrected.

\section{Proposed Method}

\subsection{Overview}
CHORD decouples proposing candidate biases from testing them for more reliable bias discovery. A bias pattern estimated from the target features alone may mix the spurious split with semantic variation, noise, or artifacts of the target geometry. We therefore propose candidates from several representations of the same samples and test each one in the target feature space. Discovery can thus draw on multiple sources and reduce the impact of noise from any single source, while the test stays tied to the model being corrected.

\begin{figure*}[!t]
    \centering
    \includegraphics[width=0.9\linewidth]{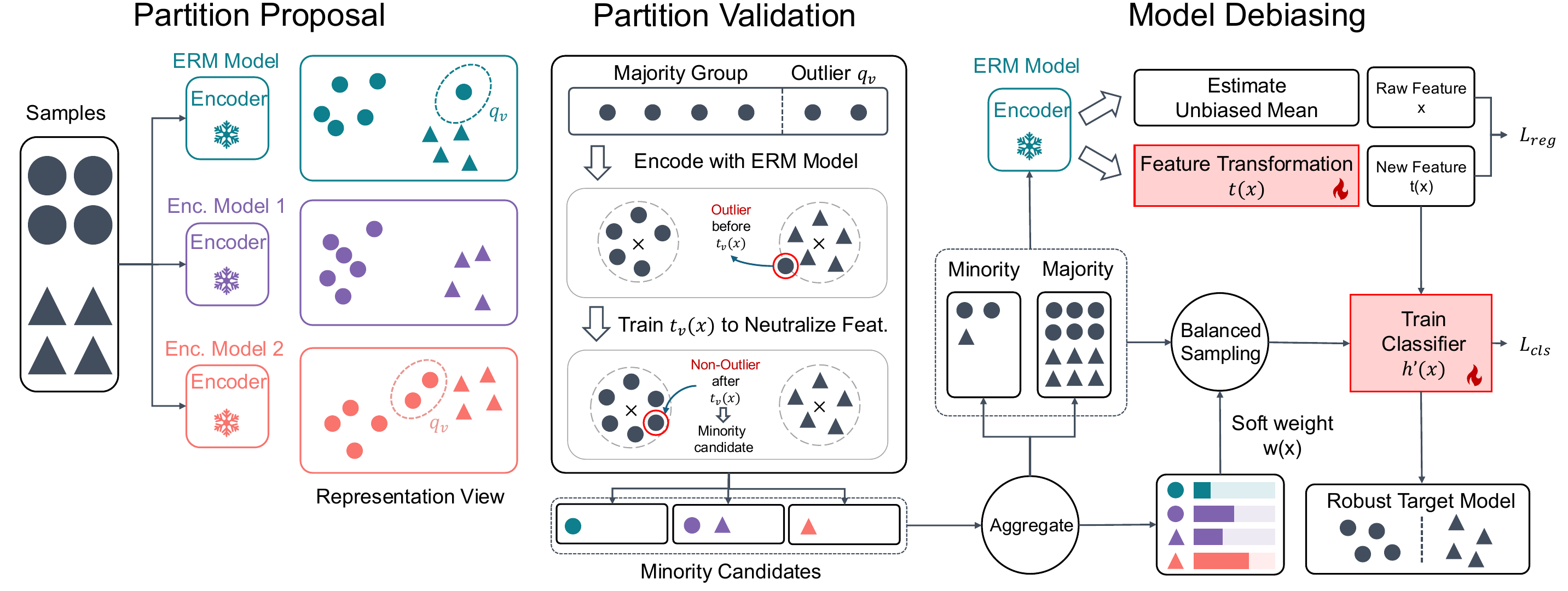}
    \caption{Overview of CHORD. CHORD separates proposing biases from testing them for reliable bias discovery in the target model. First, each view proposes a candidate bias. The intuition is that a noisy candidate bias pattern in one representation may be denoised in another. Each candidate partitions samples into majority and outlier subsets. We then test the candidate in the target space by removing the feature difference between the two subsets. Removing a real bias improves within-class compactness by moving outliers closer to their class centroid. Outliers whose nearest-centroid assignments return to their labeled class are used as minority candidates. Finally, we aggregate this evidence across views to retrain the classifier head.}\label{fig:pipe}
\end{figure*}

\subsection{Problem Formulation}
Let $\mathcal{D}=\{(x_i,y_i)\}_{i=1}^{N}$ be a set drawn from a biased distribution, with $y_i\in Y=\{1,\dots,K\}$ and $\mathcal{D}_k$ the samples of class $k$. For analysis, each example belongs to a latent group $g_i=(a_i,y_i)\in G=A\times Y$, where $a_i$ is a spurious attribute correlated with the label; neither $a_i$ nor $g_i$ is available during training. 
We start from an ERM model with encoder $f_0$ and classifier
head $h$, trained to minimize the average classification
loss on the training set. 
Because this average objective is dominated by majority groups, strong mean accuracy can mask poor minority-group generalization \cite{geirhos2020shortcut,tsirigotis2024group}. We therefore evaluate robustness by worst-group accuracy (WGA), $\operatorname{Acc}_{\mathrm{wg}}(h)=\min_{g\in G}\operatorname{Acc}_g(h)$, where $\operatorname{Acc}_g(h)$ is the accuracy on group $g$.
We denote the target feature by $z_i=f_0(x_i)\in\mathbb{R}^{D}$
and keep $f_0$ frozen throughout debiasing.
CHORD starts from the premise suggested by DFR \cite{kirichenko2022last}: even when an ERM classifier relies on a spurious cue, its frozen representation often retains sufficient core features for the task. Debiasing can therefore use the learned features rather than retraining the encoder, provided that the core signal remains recoverable. 

\subsection{Multi-View Partition Proposal}
\label{sec:partition}
Under the sampling-bias setting widely used in group robustness \cite{sagawa2020investigation,deng2024robust,chen2024understanding}, non-representative sampling can make an attribute spuriously correlate with the label \cite{an2021why}. Each class is then dominated by a majority pattern, while a smaller set of bias-conflicting samples violates that pattern \cite{torralba2011unbiased}. CHORD therefore expresses a candidate bias as a within-class partition, a hypothesized split of each class into two subsets that follow different patterns. Such a partition may also reflect other within-class variation, motivating the target-space test in \cref{sec:transform}. Because a partition is only an assignment of samples, it can be proposed in any feature space and tested in the target space. A view is a feature space in which such a partition is proposed. Its only role is to propose a partition of the labeled samples. CHORD consider three types of views, the target view $v_0$, pretrained views $\mathcal{V}_{\mathrm{pre}}$, and a geometry-enhanced view $v_{geo}$. Each view $v$ assigns every sample a feature $z_i^{(v)}$ and each class $k$ a centroid $\mu_k^{(v)}=\frac{1}{|\mathcal{D}_k|}\sum_{i:y_i=k}z_i^{(v)}$. All views observe exactly the same labeled samples, so their partitions are comparable sample by sample. All encoders stay frozen. Non-target views only propose partitions and are discarded after aggregation. They play no part in fitting transformations, training the classifier, or inference. We further explore constructing views from heterogeneous modalities in the Discussion.

\textbf{Target view $v_0$.}
The target view is the representation of the model being debiased, $z_i^{(0)}=z_i$ and $\mu_k^{(0)}=\mu_k$. It is also the space in which every proposed partition is tested. Using this view alone recovers the single-view setting.

\textbf{Pretrained views $\mathcal{V}_{\mathrm{pre}}$.}
Each pretrained view uses a frozen encoder $f_v$ trained with a different objective and training data, such as self-supervised or vision--language pretraining. For each $v\in\mathcal{V}_{\mathrm{pre}}$, the encoder produces features $z_i^{(v)}=f_v(x_i)$. The encoders used for each modality are listed in \cref{sec:exp}. Because these features are not shaped by fitting the target labels, their within-class structure can expose differences that are obscured in the target view.

\textbf{Geometry-enhanced view $v_{geo}$.}
To obtain an additional feature space, we modify the geometry of the target features. We compute the within-class scatter
\begin{equation}
    S_w=\frac{1}{N-K}\sum_{i=1}^{N}(z_i-\mu_{y_i})(z_i-\mu_{y_i})^\top=Q\Lambda Q^\top,
    \label{equ:scatter} 
\end{equation}
and define the shrinkage-whitened feature $z_i^{(\mathrm{geo})}=(\Lambda+\rho\bar\lambda I)^{-1/2}Q^\top(z_i-\bar z),$
where $\bar z$ is the global mean, $\bar\lambda$ is the mean eigenvalue, and $\rho$ is a constant. Whitening down-weights within-class directions of high variance and up-weights those of low variance, so a difference dominated by large semantic variation in the original metric can become prominent. 
    
\paragraph{Relative-distance partitioning.}
We here propose candidate bias as a within-class partition using relative prototype distance. A spurious attribute divides each class into bias-aligning samples that follow the majority pattern and bias-conflicting samples that do not. And this drives one of the groups lie closer to a competing centroid, enabling us to split each class by this criterion. Every view proposes a partition with the same rule. In view $v$, we compute each sample's distance relative to the nearest competing class $d_i^{(v)}=\|z_i^{(v)}-\mu_{y_i}^{(v)}\|_2^2-
    \min_{u\neq y_i}\|z_i^{(v)}-\mu_u^{(v)}\|_2^2.$
A positive value indicates that the sample is closer to a competing class centroid than to the centroid of its labeled class. We accordingly divide each class into $\mathcal{O}_k^{(v)}=\{i\mid y_i=k,\ d_i^{(v)}>0\}$ and $\mathcal{I}_k^{(v)}=\{i\mid y_i=k,\ d_i^{(v)}\leq0\}$. The collection $q_v=\{(\mathcal{I}_k^{(v)},\mathcal{O}_k^{(v)})\}_{k=1}^{K}$ defines the partition proposed by view $v$, with proposed outliers $\mathcal{O}^{(v)}=\bigcup_k\mathcal{O}_k^{(v)}$. Only this sample assignment is carried forward; mean estimation and transformation take place in the target space.

\subsection{Partition Validation}
\label{sec:transform}
We then transfer partitions to the target feature space to assess whether they capture bias-related variation. The central observation is that if a meaningful partition $q_v$ captures variation caused by spurious correlations, suppressing it increases within-class compactness while leaving the class-consistent core features intact. To limit the capacity to fit uninformative partitions, we constrain the mapping to an affine transformation with a diagonal linear component and parameters shared across all samples and classes. 
Let $\bar z_S$ denote the mean target feature over subset $S$. For the partition $q_v$, we estimate the class-wise neutralized mean $\tilde\mu_k^{(v)}=(\bar z_{\mathcal{O}_k^{(v)}}+\bar z_{\mathcal{I}_k^{(v)}})/2$. The balanced mean removes the effect of unequal subset frequencies from the fitting target.  When a class has no proposed outliers, we use its original prototype.
For each partition, we learn a shared lightweight diagonal transformation in the target feature space, $t_v(z)=s_v\odot(z-c_v)+c_v$,
by minimizing
\begin{equation}
    \mathcal{L}_{t}^{(v)}=
    \frac{\lambda}{D}\sum_{d=1}^{D}s_{v,d}+
    \frac{1}{N}\sum_{i=1}^{N}
    \left\|t_v(\mathbf{sg}[z_i])-\tilde\mu_{y_i}^{(v)}\right\|_2,
    \label{equ:reg}
\end{equation}
where $\lambda$ scales the penalty and $\mathbf{sg}[\cdot]$ denotes stop-gradient. 

For each class, the difference between the two subset means after transformation is $s_v\odot(\bar z_{\mathcal{O}_k^{(v)}}-\bar z_{\mathcal{I}_k^{(v)}})$. 
Attenuating a channel therefore suppresses the partition-related difference. Both subsets of each class are fitted to the same neutralized mean, encouraging the removal of attribute differences that a classifier could otherwise exploit. Fitting different classes to their own neutralized mean discourages the loss of class-discriminative information.
Samples displaced toward another class by the spurious attribute should therefore move back toward their labeled class after transformation. We identify reversal samples among the target outliers whose nearest-centroid assignments change back to their labeled class:
$\mathcal{R}^{(v)}=\{i\in\mathcal{O}^{(v)}\mid
i\in\mathcal{O}^{(0)},\
\arg\min_k\|t_v(z_i)-\tilde\mu_k^{(v)}\|_2=y_i\}$.
The first condition restricts candidates to target outliers because
auxiliary-view deviations may be irrelevant to the target representation.
The second requires reversal as evidence that neutralization can correct
these deviations.
These reversal samples serve as minority candidates for classifier retraining. A partition can thus provide useful correction evidence without exactly recovering the underlying bias groups. A partition that produces no reversal samples contributes no sampling weight. Across views, selection and weighting operate within the target-outlier set $\mathcal{O}^{(0)}$.

\subsection{Reversal Sample Aggregation}
\label{sec:aggregation}
\paragraph{View admission.}

Target and pretrained views contribute their nonempty reversal sets. Let $\mathcal{R}^{(\mathrm{pre})}=\bigcup_{v\in\mathcal V_{\rm pre}}\mathcal{R}^{(v)}$,
$r=|\mathcal{R}^{(\mathrm{geo})}|$, and
$b=|\mathcal{O}^{(\mathrm{geo})}\setminus\mathcal{O}^{(0)}|$.
Here, $b$ counts geometry-proposed outliers that are correctly
classified by the original target prototypes. We use a heuristic admission rule for the geometry-enhanced view, admitting it only if the fifth percentile of $\operatorname{Beta}(r+1,b+1)$ exceeds $0.5$ and $|\mathcal{R}^{(\mathrm{geo})}\setminus(\mathcal{R}^{(0)}\cup\mathcal{R}^{(\mathrm{pre})})|
>
|\mathcal{R}^{(\mathrm{pre})}\setminus\mathcal{R}^{(0)}|.$
The first condition favors corrected target outliers over disagreements on target-consistent samples. Beta smoothing and a lower quantile temper small-count evidence; this is a heuristic score, not a calibrated bias probability. The second requires geo to add more evidence than pretrained views add beyond the target. Neither condition uses group annotations or evaluation scores. 

Let $\mathcal{V}_{+}$ contain the admitted views with nonempty reversal sets.
For each view, 
$\mathcal{R}_k^{(v)}=\{i\in\mathcal{R}^{(v)} \mid y_i=k\}$
and $K_v=|\{k \mid \mathcal{R}_k^{(v)}\neq\emptyset\}|$.
We assign equal total weight to each supported class within a view,
then average the normalized weights across views, \mbox{$w_i=(1/|\mathcal{V}_{+}|)\sum_{v\in\mathcal{V}_{+}:\,i\in\mathcal{R}^{(v)}}1/(K_v\,|\mathcal{R}_{y_i}^{(v)}|)$}.

The resulting $w_i$ aggregates the minority-candidate evidence received by sample $x_i$. Normalization accounts for supported classes and reversal-set sizes; samples unique to one view can still contribute. A larger value causes the sample to participate more frequently as a member of the minority group during classifier updating. We do not select a single best view or convert $w_i$ into a hard group annotation. All admitted partitions contribute through the reversal samples they produce. The auxiliary encoders and their transformations are discarded after $w_i$ has been computed.

\subsection{Target Model Update}
A classifier’s reliance on spurious correlations is shaped by both its input features and how samples are weighted during training. CHORD addresses bias by training a new classifier using two signals. First, the target transformation $t_0$ learned from the target partition $q_0$ provides the transformed feature $t_0(z_i)$. Second, the aggregated weight $w_i$ determines the sampling frequency within the minority group $\mathcal{R}=\bigcup_{v\in\mathcal{V}_{+}}\mathcal{R}^{(v)}$. Thus, learning the new classifier does not require the original classifier’s predictions.
We define the majority group as $\mathcal{M}=\{i\mid d_i^{(0)}\leq0\}$. Each balanced batch contains one subset drawn from $\mathcal{R}$ according to $w_i$ and an equally sized subset drawn uniformly from $\mathcal{M}$. Importantly, samples from both groups are represented by the same transformed target feature $t_0(z)$.

Let $(x_i,y_i)$ denote a minority sample drawn according to $w_i$ and $(x_j,y_j)$ a majority sample drawn from $\mathcal{M}$. We update only the classifier head $h'$ using
\begin{equation}
\begin{split}
    \mathcal{L}_{\mathrm{cls}}={}&
    \mathcal{L}_{\mathrm{CE}}\left(h'(\mathbf{sg}[t_0(z_i)]),y_i\right)\\
    &+\mathcal{L}_{\mathrm{CE}}\left(h'(\mathbf{sg}[t_0(z_j)]),y_j\right).
\end{split}
\end{equation}
The transformation reduces the within-class variation associated with the spurious attribute, while the sampling weight prevents the classifier head from remaining dominated by the original majority pattern. At inference, CHORD retains only the target encoder $f_0$, target transformation $t_0$, and updated classifier head $h'$. The auxiliary encoders and transformations $\{t_v\}_{v\neq0}$ are used only during training.

\begin{table*}[!t]
    \caption{Mean accuracy (mAcc) and worst-group accuracy (WGA) on four widely used benchmarks. For CHORD, we report means over 10 classifier seeds with fixed ERM features, with the standard deviations shown for WGA.}

    \centering
    \setlength{\tabcolsep}{2.5mm}
    \small{
    \resizebox{0.9\textwidth}{!}{\begin{tabular}{lccccccccccc}
    \toprule
    &\multicolumn{2}{c}{\textsc{Labels}} & \multicolumn{2}{c}{\textsc{Waterbirds}} & \multicolumn{2}{c}{\textsc{CelebA}} & \multicolumn{2}{c}{\textsc{MultiNLI}} & \multicolumn{2}{c}{\textsc{CivilComments}} & \textsc{Mean} \\
    \cmidrule(r{4pt}){2-3} \cmidrule(r{4pt}){4-5} \cmidrule(r{4pt}){6-7} \cmidrule(r{4pt}){8-9} \cmidrule(r{4pt}){10-11} \cmidrule(r{4pt}){12-12}
    & \textsc{Tr.} & \textsc{Val} & \textsc{mAcc} & WGA & \textsc{mAcc} & WGA& \textsc{mAcc} & WGA & \textsc{mAcc} & WGA  & WGA \\
    \midrule

    \textsc{ERM} & \xmark & \xmark & $97.30$ & $72.60$ & $95.60$ & $47.20$ & $82.09$ & $68.12$ & $92.34$ & $57.06$&61.25\\\midrule
     \textsc{JTT\cite{liu2021just}}& \xmark & \cmark &93.30&86.70&{88.00}& 81.10& $78.60$ & 72.60 & {91.10} & 69.30 &77.43 \\
          \textsc{MT\cite{asgari2022masktune}}& \xmark & \xmark & $93.00$ & $86.40$ & $91.30$ & $78.00$ &\multicolumn{5}{c}{------------ Not Applicable ------------} \\

\textsc{CNC\cite{zhang2022correct}} & \xmark & \xmark &90.90&{88.50}&{89.90}& 88.80& - & - & 81.70 & 68.90 & - \\
\textsc{AFR\cite{qiu2023simple}}& \xmark & \cmark &94.20&{90.40}&{91.30}& {82.00}& {81.40} & 73.40 & 89.80 & 68.70  & 78.63 \\
      \textsc{EA\cite{ye2025evidential}} & \xmark & \xmark & 95.10 & 92.20 & 92.30 & 84.40 & 80.60 & 74.50 & 87.10 & 80.20 & 82.83 \\
      \textsc{SP\cite{zheng2025shortcutprobe}} & \xmark & \xmark & 95.00 & 90.80 & 91.40 & 83.40 & 82.60 & 74.30 & 88.50 & 79.90 & 82.10 \\

      \textsc{NSF\cite{li2025nsf}}  & \xmark & \xmark & 95.65 & 91.12 & 88.70 & 84.27 & 80.43 & 73.12 & 87.19 & 79.51 & 82.01 \\
      \textsc{\textbf{CHORD}} & \xmark & \xmark & 95.48 & $\textbf{93.11}_{\pm 0.44}$ & 90.78 & $\textbf{89.06}_{\pm 0.50}$ & 81.40 & $\textbf{75.25}_{\pm 0.11}$ & 88.18 & $\textbf{80.52}_{\pm 0.30}$ & \textbf{84.49} \\
      \midrule
      \multicolumn{12}{c}{\textit{Methods using bias/group supervision}} \\
      \midrule
       \textsc{GDRO\cite{sagawa2019distributionally}}& \cmark & \cmark & $93.50$ & $91.40$ & $92.90$ & $88.90$ & 81.40& 77.70& 88.90& 69.90 &81.98\\\textsc{DFR\cite{kirichenko2022last}} & \xmark & \cmark &94.20&92.90&91.30& 88.30& 82.10 & 74.70 & 87.20 & 70.10 & 81.50\\
              \textsc{uLA\cite{tsirigotis2024group}} & \xmark & \cmark & $91.50$ & $86.10$ & ${93.90}$ & $86.50$&\multicolumn{5}{c}{------------ Not Applicable ------------}\\
    \bottomrule
    \end{tabular}
    }
    }
     \label{tab:results-celeba}
\end{table*}

\begin{table}[!t]
    \centering
    \caption{Macro-average of mAcc and WGA on CREMA-D and MEAD. GDRO$^{*}$ averages attribute-matched experts, requiring oracle model switching.}
    \label{tab:speech_results}
    \setlength{\tabcolsep}{8pt}
    \resizebox{0.9\columnwidth}{!}{\begin{tabular}{lcccccc}
    \toprule
    & \multicolumn{2}{c}{CREMA-D} & \multicolumn{2}{c}{MEAD} & \multicolumn{2}{c}{Mean} \\
    \cmidrule(lr){2-3} \cmidrule(lr){4-5} \cmidrule(lr){6-7}
    \textsc{Method} & mAcc & WGA & mAcc & WGA & mAcc & WGA \\
    \midrule
    ERM   & 77.08 & 59.63 & 60.51 & 48.35 & 68.80 &	53.99 \\
    NSF   & 76.99 & 60.63  & 61.67 & 50.77 & 69.33 & 55.70 \\
    CHORD & 77.11 &	\textbf{67.02} & 61.19 & \textbf{52.84} & 69.15 & \textbf{59.93} \\
    \midrule
    GDRO$^{*}$~\cite{sagawa2019distributionally} & 76.02 & 59.20
    & 61.30 & 47.29
    & 68.67 & 53.25  \\
    \bottomrule
    \end{tabular}
    }
    \end{table}

\section{Experiments}
\label{sec:exp}

\subsection{Experimental Settings}

\textbf{Datasets.} We evaluate CHORD on three tasks.

\emph{Vision.}
Waterbirds \cite{sagawa2019distributionally} contains 4,795 landbird and waterbird images with spuriously correlated backgrounds.
CelebA \cite{liu2015deep} contains over 200,000 face images; we predict hair color and define groups by gender.
CheXpert \cite{irvin2019chexpert} contains over 200,000 chest radiographs; we adopt the binary task and group construction of SubPopBench \cite{yang2023change}.

\emph{Language.}
MultiNLI \cite{williams2018broad} contains 412,349 labeled sentence pairs labeled as entailment, contradiction, and neutral; negation terms is the spurious attribute.
CivilComments-WILDS \cite{koh2021wilds} contains 448,000 comments labeled as toxic or non-toxic, with groups defined by demographic identity.

\emph{Speech.}
For emotion recognition, CREMA-D \cite{cao2014cremad} provides 7,442 recordings from 91 actors across six emotions.
We evaluate five attributes: sex, binary race, median-split age, age quartiles, and the intersection of sex and binary race.
Sex, binary race, and binary age each define 12 emotion–attribute groups; age quartiles and sex–race intersections each define 24 groups.
For MEAD \cite{kaisiyuan2020mead}, we use a fixed subset of 31,729 recordings from 48 actors across eight emotions. The actor-disjoint training, validation, and test splits contain 21,125, 5,302, and 5,302 recordings from 32, 8, and 8 actors, respectively. We evaluate groups defined by emotion and either speaker sex or emotional intensity.

\textbf{Metrics.}
We report test mean accuracy (mAcc) and worst-group accuracy (WGA). For CREMA-D and MEAD, we average the attribute-specific WGAs over five and two group definitions, respectively; individual and intersectional results appear in the multi-attribute analysis. All CHORD results use the last-epoch checkpoint.

\textbf{Network Architecture.}
Image tasks use ImageNet-pretrained ResNet-50 \cite{he2016deep}; Waterbirds and CelebA additionally use ViT-B/16 \cite{dosovitskiy2021vit} for cross-architecture evaluation.
Language tasks use BERT-base-uncased \cite{devlin2019bert}, and speech tasks use fine-tuned wav2vec~2.0-base \cite{baevski2020wav2vec2}.

\textbf{Implementation Details.}
All experiments use PyTorch \cite{paszke2019pytorch} on an Nvidia GeForce 3090 GPU.
We train ERM following DFR \cite{kirichenko2022last} and use SubPopBench \cite{yang2023change} preprocessing and groups for CheXpert.
After ERM training, partition proposals, centroid estimates, transformation fitting, evidence selection and aggregation, and classifier retraining all use the validation split with no group labels. CHORD use AdamW with a lr of 1e-3 and 0 weight decay.
We set $\lambda=10$ and $\rho=0.1$. CHORD uses 10 classifier seeds per fixed ERM model, measuring classifier-training variability rather than independent encoder training.
Due to space constraints, we refer readers to the code for further details.

\textbf{Baseline protocol.}
To avoid overstating our gains, we report the better of the published results and our reproductions for each competing method. Our speech GDRO result use one model seed per attribute without hyperparameter tuning; tuning may further improve performance. 

\textbf{Pretrained Representation Views.}
All auxiliary encoders are frozen and used only during training.
ResNet-50 image tasks use DINO, SimCLR, Barlow Twins, SwAV, CLIP, and BLIP; ViT-target experiments and CheXpert use DINO, MAE, and CLIP.
Language uses BERT, RoBERTa, ELECTRA, SimCSE, CLIP, and BLIP; speech uses wav2vec~2.0, HuBERT, WavLM, and Whisper.

\subsection{Comparison with State-of-the-Art Methods}

We compare against JTT \cite{liu2021just}, MT \cite{asgari2022masktune}, CNC \cite{zhang2022correct}, AFR \cite{qiu2023simple}, EA \cite{ye2025evidential}, SP \cite{zheng2025shortcutprobe}, GDRO \cite{sagawa2019distributionally}, DFR \cite{kirichenko2022last}, NSF\cite{li2025nsf}, and uLA \cite{tsirigotis2024group} on the four popular benchmarks under the same group definitions.
As shown in \cref{tab:results-celeba}, CHORD obtains the highest WGA among methods without bias or group supervision on all four benchmarks, increasing their mean WGA from 82.01\% for NSF to 84.49\%. We report both mAcc and WGA to expose their tradeoff.

\subsection{Results on Speech Emotion Recognition}
\label{sec:speech_results}

We evaluate the method on CREMA-D and MEAD using wav2vec~2.0.
As shown in \cref{tab:speech_results}, CHORD improves WGA over the
paired ERM models by 7.39 on CREMA-D and 4.49 on MEAD.
For reference, GDRO$^{*}$ trains one expert per evaluation
attribute, five on CREMA-D and two on MEAD.
We average each expert's mAcc
and its WGA under the matched evaluation attribute.
GDRO$^{*}$ chooses a different expert for each evaluation attribute. We therefore report it as an attribute-matched oracle reference, not as the performance of a single model under all attribute definitions.
Under this, CHORD exceeds GDRO$^{*}$ by 7.82 in WGA on CREMA-D and 5.55 on MEAD, using a single model trained without attribute annotations.

\subsection{Mitigating Bias in Medical Domain}
We conduct experiments on the CheXpert \cite{irvin2019chexpert} dataset to validate the proposed method in medical imaging. As shown in \cref{tab:med}, CHORD obtains a WGA of 71.43\%.

\subsection{Ablation Study}
\label{sec:ablations}

\begin{table}[!t]
    \centering

    \caption{Results on CheXpert. Both mAcc and WGA are reported; group-supervised GDRO is included as a reference. \label{tab:med}}
    \smallskip\noindent
    \setlength{\tabcolsep}{9pt}
    \resizebox{0.8\columnwidth}{!}{\begin{tabular}{cccccc}
    \toprule
     & \multicolumn{2}{c}{\textsc{Bias Label}} & \multicolumn{3}{c}{\textsc{Chexpert}} \\
     \cmidrule(r{4pt}){2-3} \cmidrule(r{4pt}){4-6}
     & \textsc{Train} & \textsc{Val} & \textsc{mAcc} & \textsc{WGA} & \textsc{Gain} \\ \midrule
    \textsc{ERM} & \xmark & \xmark & 89.78 & 31.21 & - \\
    \textsc{NSF} & \xmark & \xmark & 81.94 & 70.21 & +39.00 \\
    \textsc{CHORD} & \xmark & \xmark & 81.62 & \textbf{71.43} & \textbf{+40.22} \\ \midrule
    \textsc{GDRO\cite{sagawa2019distributionally}} & \cmark & \cmark & 78.90 & 74.50 & +43.29 \\ \bottomrule
    \end{tabular}}
    \end{table}

    \begin{table}[!tbp]
        \centering
        \caption{Component ablation on CelebA. MV adds multiple reversal-sample sets, SW aggregates their support as soft sampling weights, and MD uses the target transformation $t_0$ for the final classifier update. We report mean WGA over ten random seeds.}
        \label{tab:celeba_component_ablation}
        \setlength{\tabcolsep}{12pt}
        \resizebox{0.9\columnwidth}{!}{\begin{tabular}{lcccc}
        \toprule
        \textsc{Method} & MV & SW & MD & WGA \\
        \midrule
        ERM       &  &  &  & 47.20 \\
        Single View &  &  & \cmark & $84.27$ \\
        MV-Hard   & \cmark &  & \cmark & $86.78$ \\
        CHORD     & \cmark & \cmark & \cmark & $\mathbf{89.06}$ \\
        \bottomrule
        \end{tabular}
        }
        \end{table}

\textbf{Pipeline Components.}
We first examine the three operations used in CHORD in \cref{tab:celeba_component_ablation}.
MV constructs multiple reversal-sample sets, SW retains their cross-view support as soft sampling weights, and MD uses the target transformation $t_0$ when updating the final classifier.
Because SW requires multiple sets to aggregate, it is evaluated only with MV enabled.
Replacing SW with a hard union reduces WGA from 89.06\% to 86.78\%. This supports the combined weighting rule, accounting for view support, supported classes, and reversal-set size, over treating all candidates in the union equally.

\textbf{Sources of Multi-View Evidence.}
We next separate pretrained representation views from the geometry-enhanced view in \cref{tab:multinli_framework_ablation}.
Representation views improve the single-view configuration from 73.12\% to 74.19\% WGA, whereas the geometry-enhanced view alone reaches 73.36\%.
Combining both sources yields the best WGA of 75.25\%.
The geometry-enhanced partition therefore provides complementary evidence when the pretrained views expose highly similar within-class structures, rather than serving as a substitute for representation diversity.

\begin{table}[!t]
    \centering
    \caption{Effect of individual pretrained views on CelebA. We report WGA after adding one view to Target Only in (a) or removing one from CHORD in (b). Every addition improves WGA, while most removals lower it. BT denotes Barlow Twins.}
    \label{tab:celeba_view_contribution}
    \setlength{\tabcolsep}{1.5pt}
    \resizebox{0.9\columnwidth}{!}{\begin{minipage}[t]{0.48\columnwidth}
    \centering
    \textbf{(a) Add one view}\par
    \smallskip
    \small
    \begin{tabular}{@{}lc@{\hspace{6pt}}lc@{}}
    \toprule
    \multicolumn{4}{c}{Target Only: 84.27} \\
    \midrule
    \textsc{Conf.} & WGA & \textsc{Conf.} & WGA \\
    \cmidrule(r{3pt}){1-2}\cmidrule(l{3pt}){3-4}
    BT   & 86.32 & DINO   & 87.39 \\
    BLIP & 87.49 & SimCLR & 87.78 \\
    CLIP & 86.84 & SwAV   & 86.71 \\
    \bottomrule
    \end{tabular}
    \end{minipage}
    \hfill
    \begin{minipage}[t]{0.48\columnwidth}
    \centering
    \textbf{(b) Remove one view}\par
    \smallskip
    \small
    \begin{tabular}{@{}lc@{\hspace{6pt}}lc@{}}
    \toprule
    \multicolumn{4}{c}{CHORD: \textbf{89.06}} \\
    \midrule
    \textsc{Conf.} & WGA & \textsc{Conf.} & WGA \\
    \cmidrule(r{3pt}){1-2}\cmidrule(l{3pt}){3-4}
    BT   & 86.86 & DINO   & 88.94 \\
    BLIP & 88.33 & SimCLR & 87.66 \\
    CLIP & 87.82 & SwAV   & 89.12 \\
    \bottomrule
    \end{tabular}
    \end{minipage}
    }
\end{table}
\textbf{Individual Representation Views.}
Finally, \cref{tab:celeba_view_contribution} examines whether the gain depends on a particular pretrained encoder with an ERM-trained ResNet-50.
Each target-plus-one-view configuration exceeds the Single View reference.
Most leave-one-out results fall below the full CHORD reference; removing SwAV gives 89.12\% versus 89.06\%, so adding a view is not uniformly beneficial.
\begin{table}[!tbp]
    \centering
    \caption{View-level ablation on MultiNLI. }
    \label{tab:multinli_framework_ablation}
    \setlength{\tabcolsep}{3.5pt}
    \resizebox{0.9\columnwidth}{!}{\begin{tabular}{lcccc}
    \toprule
    \textsc{Method} & Target View & Rep. Views & Geo. View & WGA \\
    \midrule
    ERM                         &        &        &        & 68.12 \\
    Target Only                 & \cmark &        &        & 73.12 \\
    Target + Pre. Views  & \cmark & \cmark &        & 74.19 \\
    Target + Geo. View         & \cmark &        & \cmark & 73.36 \\
    CHORD                       & \cmark & \cmark & \cmark & \textbf{75.25}\\
    \bottomrule
    \end{tabular}
    }
    \end{table}

\subsection{Cross-Architecture Applicability}
\label{sec:cross_arch}

We further examine whether the method depends on a particular target architecture.
\Cref{tab:cross_arch} compares ERM and CHORD using both ResNet-50 and ViT-B/16 on Waterbirds and CelebA.
These results show that CHORD improves worst-group accuracy over ERM with both architectures.
\begin{table}[!tbp]
\centering
\caption{Cross-architecture comparison on Waterbirds and CelebA. We report mean accuracy (mAcc) and worst-group accuracy (WGA) for ResNet-50 and ViT-B/16. CHORD improves WGA in all four dataset--architecture settings.}
\label{tab:cross_arch}
\setlength{\tabcolsep}{8pt}
\resizebox{0.9\columnwidth}{!}{\begin{tabular}{llcccc}
\toprule
& & \multicolumn{2}{c}{ResNet-50} & \multicolumn{2}{c}{ViT-B/16} \\
\cmidrule(lr){3-4} \cmidrule(lr){5-6}
\textsc{Dataset} & \textsc{Method} & mAcc & WGA & mAcc & WGA \\
\midrule
\multirow{2}{*}{Waterbirds}
& ERM   & 97.30 & 72.60 & 92.89 & 78.97 \\
& CHORD & {95.48} & \textbf{93.11} & 96.64 & \textbf{89.42} \\
\midrule
\multirow{2}{*}{CelebA}
& ERM   & 95.60 & 47.20 & 96.21 & 39.72 \\
& CHORD & 90.78 & \textbf{89.06} & 93.26 & \textbf{87.36} \\
\bottomrule
\end{tabular}
}
\end{table}

\subsection{Robustness Across Attribute Definitions}
\label{sec:attribute_robustness}

\textbf{Multiple Post-Hoc Attributes}
CREMA-D allows us to evaluate the same trained model under five post-hoc group definitions covering sex, race, two age discretizations, and the intersection of sex and race.
As shown in \cref{tab:cremad_attributes}, CHORD improves WGA over ERM under all five definitions, although none of these attributes is used by the method during training.
Against ERM, the largest improvement occurs under age-quartile
grouping, where CHORD raises WGA by 11.89.
These results support simultaneous robustness to multiple possible group definitions, rather than the stronger claim that each latent bias is separately recovered or assigned to a particular view.
We also evaluate five GDRO models trained with different group annotations for reference, each under all five evaluation attributes.

\begin{table}[!tbp]
\centering
\caption{Multi-attribute evaluation on CREMA-D (\%). Each row evaluates one model across five post-hoc group definitions. Train attr. specifies GDRO's group annotation. Underlined entries are the attribute-matched GDRO results. CHORD results are averaged over 10 classifier seeds on fixed ERM features; GDRO uses one model seed per attribute. Age-B/Q denote binary/quartile age groups; SexRace denotes sex $\times$ race. CHORD achieves highest WGA under all five definitions without attribute-specific training.}
\label{tab:cremad_attributes}
\setlength{\tabcolsep}{2.5pt}
\resizebox{0.9\columnwidth}{!}{\begin{tabular}{llccccccc}
\toprule
\textsc{Method} & Train attr. & mAcc & \multicolumn{6}{c}{WGA by evaluation attribute} \\
\cmidrule(lr){4-9}
& & & Sex & Race & Age-B & Age-Q & SexRace & Mean \\
\midrule
ERM   & -- & 77.08	& 62.00	& 61.54	& 62.50	& 52.63	& 59.46	& 59.63\\
CHORD & -- & \textbf{77.11}	& \textbf{66.70}	& \textbf{69.62}	& \textbf{69.41}	& \textbf{64.52}	& \textbf{64.86}	& \textbf{67.02}\\
\midrule
\multirow{5}{*}{GDRO}
& Sex     & 75.92 & \underline{59.41} & 62.32 & 61.32 & 55.10 & 59.38 & 59.51 \\
& Race    & 74.04 & 55.56 & \underline{57.89} & 60.49 & 53.49 & 48.44 & 55.17 \\
& Age-B   & 77.08 & 63.37 & 63.08 & \underline{64.15} & 55.56 & 56.67 & 60.57 \\
& Age-Q   & 76.54 & 58.59 & 62.32 & 65.26 & \underline{57.89} & 51.56 & 59.12 \\
& SexRace & 76.54 & 60.40 & 62.32 & 62.50 & 57.14 & \underline{56.67} & 59.81 \\
\bottomrule
\end{tabular}
}
\end{table}

\textbf{Exchanging Target and Evaluation Attributes}
The preceding analysis evaluates multiple possible spurious attributes for a fixed prediction target.
As a complementary test, we exchange Blond Hair and Gender between the target and evaluation attributes on CelebA.
We keep the standard CelebA CHORD configuration for both assignments without task-specific retuning.
As shown in \cref{tab:celeba_attribute_swap}, CHORD attains 89.06\% WGA for Blond Hair prediction and 96.03\% for Gender prediction, without accessing the evaluation attribute during training.
For the exchanged task, CHORD improves WGA over its matching ERM checkpoint by 8.81.
This test evaluates transfer across attribute roles; it does not imply that the two attributes are simultaneously recovered as separate latent biases.

\begin{table}[!tbp]
\centering
\caption{Attribute-role exchange on CelebA to assess transfer across prediction tasks. Blond Hair and Gender exchange their roles as the prediction target and evaluation attribute, using the standard ResNet-50 CHORD configuration without task-specific retuning. We report means over 10 random seeds, with WGA standard deviations. CHORD improves WGA under both assignments.}
\label{tab:celeba_attribute_swap}
\setlength{\tabcolsep}{3.5pt}
\resizebox{0.9\columnwidth}{!}{\begin{tabular}{llcccc}
\toprule
& & \multicolumn{2}{c}{ERM} & \multicolumn{2}{c}{CHORD} \\
\cmidrule(lr){3-4} \cmidrule(lr){5-6}
\textsc{Target} & \textsc{Eval. Attribute} & mAcc & WGA & mAcc & WGA \\
\midrule
Blond Hair & Gender     & 95.60 & 47.20 & 90.78 & $\mathbf{89.06}_{\pm 0.50}$ \\
Gender     & Blond Hair & 98.64 & 87.22 & 97.73 & $\mathbf{96.03}_{\pm 0.51}$ \\
\bottomrule
\end{tabular}
}
\end{table}

\section{Discussion}
\textbf{Transformations \& Random Views.}
On CelebA, we keep the ERM model and the final transformation fixed, averaging 10 classifier seeds and, for randomized views, 10 independently sampled partitions. Target-only achieves 82.39\% WGA; adding one random view with 50\% or 25\% within-class outliers yields 83.20\%/83.84\%. Removing the transformation from reversal checking or using original centroids yields 64.67\%/80.09\%, with the final transformation unchanged. In a separate experiment, we add one CLIP view using three partition variants: the original partition, a partition with half of its outliers replaced by inliers, and a fully randomized partition. With batch size and training steps fixed, auxiliary candidates must be target outliers; we then enable or disable the reversal check. Enabling it changes WGA from 85.90\%/84.38\%/77.10\% to 85.90\%/85.06\%/83.42\%, respectively, supporting the check's utility under corrupted partitions. Strong performance with random candidates also shows the contribution of target-space test; the small original--random gap here should be distinguished from the comparison with all pretrained partitions randomized below.

\textbf{Randomizing all pretrained partitions.}
We replace all six pretrained partitions with independent random partitions, matching each view's outlier count within each class. The target partition, final transformation, update budgets, and geo admission (off) remain fixed. On CelebA, across 10 independent randomizations. Pretrained versus random partitions yield $89.06\pm0.50\%$ versus $75.28\pm1.25\%$ WGA. Both standard deviations are across classifier seeds, with random results first averaged over randomizations per seed. 

\textbf{ERM seeds.} On CelebA, CHORD achieves $88.60\pm0.47\%$ WGA across three ERM seeds, versus $82.91\%$ for target-only; each ERM result averages 10 classifier seeds, and the standard deviation is across ERM means.

\textbf{Beyond in-modality views.} Since a view only produces a within-class partition, it need not share the target’s modality or even observe the input directly. This opens two possibilities. First, views can be built through language: 
{\small
\begin{equation}
    \begin{aligned}
    \text{Image}
&\xrightarrow{\mathrm{VLM}}\text{Caption}
\xrightarrow{\mathrm{BERT}}z_i^{(\mathrm{v})}
\xrightarrow{\text{partition}}q_{\mathrm{v}},
&
\end{aligned}
\end{equation}
}
captioning CelebA images with a VLM then encoding with BERT improves WGA by 1.5 over target-only. Because captions can be steered by prompts, this also offers a natural channel for human priors to coexist with automatic discovery. Second, views can be borrowed from another modality: 
{\small
\begin{equation}
    \begin{aligned}
    \text{Speech}
    &\xrightarrow{\text{log-Mel}}\text{Spectrogram}
    \xrightarrow{\mathrm{DINO}}z_i^{(\mathrm{v})}
    \xrightarrow{\text{partition}}q_{\mathrm{v}}.
&
\end{aligned}
\end{equation}
} 
on MEAD, rendering speech as log-Mel spectrograms and encoding them with a frozen DINOv3, raises WGA from 50.77 to 52.04 and contributes 22 reversal samples not found from the target view alone. These results suggest that CHORD can draw on representation views beyond pretrained encoders of the target modality, which also mitigates the scarcity of such encoders in specialized domains.

\textbf{Extensible partition proposals.} Replacing relative-distance partitioning with ERM errors in the target view and linear-probe (LP) errors in auxiliary views, we follow the main-experiment CHORD training recipe. CelebA WGA improves from 44.39\% with direct partition-balanced retraining to 82.33\% with target-only CHORD and 84.50\% with auxiliary proposals, demonstrating extensibility beyond the choice of representation views.

\textbf{Computation.}
On one GeForce~3090, CHORD takes 4.5/13.9\,s on Waterbirds/CelebA from cache loading to the first trained classifier while competing methods JTT and CNC require minutes to hours\cite{qiu2023simple}. Timings exclude ERM training and feature extraction; 

\textbf{Limitations.} CHORD requires recoverable core features in the frozen target representation and restricts reversal evidence to the target-outlier pool. Although CHORD is evaluated on multiple tasks, suitable pretrained encoders may be less available in specialized domains. In such settings, CHORD can still combine the target representation with its geometry-enhanced view or heterogeneous views.

\section{Conclusion}

This work introduces CHORD, a framework for mitigating spurious correlations without group annotations. CHORD separates proposing candidate biases from testing them: multiple views propose within-class partitions, and the target model tests each partition through reversal samples. This design rests on a simple observation: spurious cues are shaped by the target representation, but evidence of them can come from outside the target model. Experiments on benchmarks across vision, language, speech, and medical imaging consistently demonstrate improved worst-group accuracy.

\section*{LLM Usage Statement}
OpenAI Codex was used to assist with drafting and revising manuscript text, including the Introduction, Related Work, and Limitations, and with literature lookup and LaTeX preparation.
It also assisted with code preparation, experiment execution, and result checks.
The authors remain responsible for the scientific claims, experimental results, and final manuscript.

\end{document}